%% file: _main.tex
\input{_constants}
\arxiv 

\pdfoutput=1
\documentclass[10pt,twocolumn,letterpaper]{article}

\usepackage{makecell}
\input{cvpr_header}
\begin{document}
\title{Not All Steps are Created Equal: \\ Selective Diffusion Distillation for Image Manipulation}
\author{\authorBlock}
\maketitle

\input{00_abstract_v1}
\input{01_intro_v13}

\input{02_related_v2}

\input{03_background_v2}
\input{04_method_v7}
\input{05_experiment_v1}
\input{10_conclusion}
\input{11_acknowledgement}

{\small
\bibliographystyle{ieee_fullname}
\bibliography{11_references}
}

\ifarxiv \clearpage \input{12_appendix} \fi

\end{document}

%% file: _constants.tex
\def\cvprPaperID{5762}
\def\confName{ICCV}
\def\confYear{2023}

\def\authorBlock{
    Luozhou Wang\textsuperscript{$1,4$}\thanks{Equal contribution} \qquad
    Shuai Yang\textsuperscript{$1,4$}\footnotemark[1] \qquad
    Shu Liu\textsuperscript{$2$} \qquad
    Ying-cong Chen\textsuperscript{$1,3,4$}\thanks{Corresponding author} \quad
    \\
    \small$^1$ The Hong Kong University of Science and Technology~(Guangzhou).\quad 
    \small$^2$ SmartMore.\\
    \small$^3$ The Hong Kong University of Science and Technology.
    \small$^4$ HKUST (Guangzhou) - SmartMore Joint Lab.
}

\newif\ifreview 
\newif\ifarxiv \newcommand{\arxiv}{\arxivtrue}
\newif\ifcamera 
\newif\ifrebuttal 

%% file: cvpr_header.tex
\ifreview \usepackage[review]{cvpr} \fi
\ifarxiv \usepackage[pagenumbers]{cvpr} \fi
\ifrebuttal \usepackage[rebuttal]{cvpr} \fi
\ifcamera \usepackage{cvpr} \fi

\usepackage{graphicx}
\usepackage{amsmath}
\usepackage{amssymb}
\usepackage{booktabs}

\input{_macros}  

\usepackage{xr-hyper}

\makeatletter
\newcommand*{\addFileDependency}[1]{
  \typeout{(#1)}
  \@addtofilelist{#1}
  \IfFileExists{#1}{}{\typeout{No file #1.}}
}

\makeatother

\usepackage[pagebackref,breaklinks,colorlinks]{hyperref}
\usepackage[capitalize]{cleveref}
\crefname{section}{Sec.}{Secs.}
\crefname{table}{Table}{Tables}
\crefname{figure}{Fig.}{Figs.}

%% file: _macros.tex
\usepackage{times}
\usepackage{microtype}
\usepackage{epsfig}
\usepackage[table,xcdraw]{xcolor}
\usepackage{caption}
\usepackage{float}
\usepackage{placeins}
\usepackage{color, colortbl}
\usepackage{stfloats}
\usepackage{enumitem}
\usepackage{tabularx}
\usepackage{xstring}
\usepackage{multirow}
\usepackage{xspace}
\usepackage{url}
\usepackage{subcaption}
\usepackage{xcolor}
\usepackage[hang,flushmargin]{footmisc}
\usepackage{algorithm}
\usepackage{algpseudocode}
\usepackage{graphicx}

\ifcamera \usepackage[accsupp]{axessibility} \fi

% ...

\ifarxiv  \fi

\newcommand{\R}[1]{{%
    \textbf{%
        \ifstrequal{#1}{1}{\textcolor{red}{R#1}}{%
        \ifstrequal{#1}{2}{\textcolor{blue}{R#1}}{%
        \ifstrequal{#1}{3}{\textcolor{magenta}{R#1}}{%
        \ifstrequal{#1}{4}{\textcolor{teal}{R#1}}{%
                           \textcolor{cyan}{R#1}%
        }}}}%
    }%
}}

%% file: 00_abstract_v1.tex
\begin{abstract}
Conditional diffusion models have demonstrated impressive performance in image manipulation tasks. The general pipeline involves adding noise to the image and then denoising it. However, this method faces a trade-off problem: adding too much noise affects the fidelity of the image while adding too little affects its editability. This largely limits their practical applicability. In this paper, we propose a novel framework, Selective Diffusion Distillation (SDD), that ensures both the fidelity and editability of images. Instead of directly editing images with a diffusion model, we train a feedforward image manipulation network under the guidance of the diffusion model. Besides, we propose an effective indicator to select the semantic-related timestep to obtain the correct semantic guidance from the diffusion model. This approach successfully avoids the dilemma caused by the diffusion process. Our extensive experiments demonstrate the advantages of our framework. Code is released at \href{https://github.com/AndysonYs/Selective-Diffusion-Distillation}{https://github.com/AndysonYs/Selective-Diffusion-Distillation}.
\end{abstract}

%% file: 01_intro_v13.tex
\section{Introduction}
\label{sec:intro}
In recent years, diffusion model \cite{sohl2015deep,ho2020ddpm,song2020ddim,dhariwal2021guided,nichol2021glide,saharia2022imagen,ramesh2022dalle2,rombach2022stablediffusion,song2019score1,song2020score2} has attracted great attention in both academic and industrial communities. 
It models the Markov transition from a Gaussian distribution to a data distribution to generate high-quality images sequentially. 
The elegant formulation achieves state-of-the-art performance in various image generation benchmarks. Meanwhile, text-to-image diffusion models \cite{saharia2022imagen, rombach2022stablediffusion, nichol2021glide, ramesh2022dalle2} also demonstrate their impressive capacity in controllable image synthesis, enabling a wide range of practical applications. Among them, one of the most interesting applications is image manipulation. 

\begin{figure}[htbp]
    \centering
    \includegraphics[width=8cm]{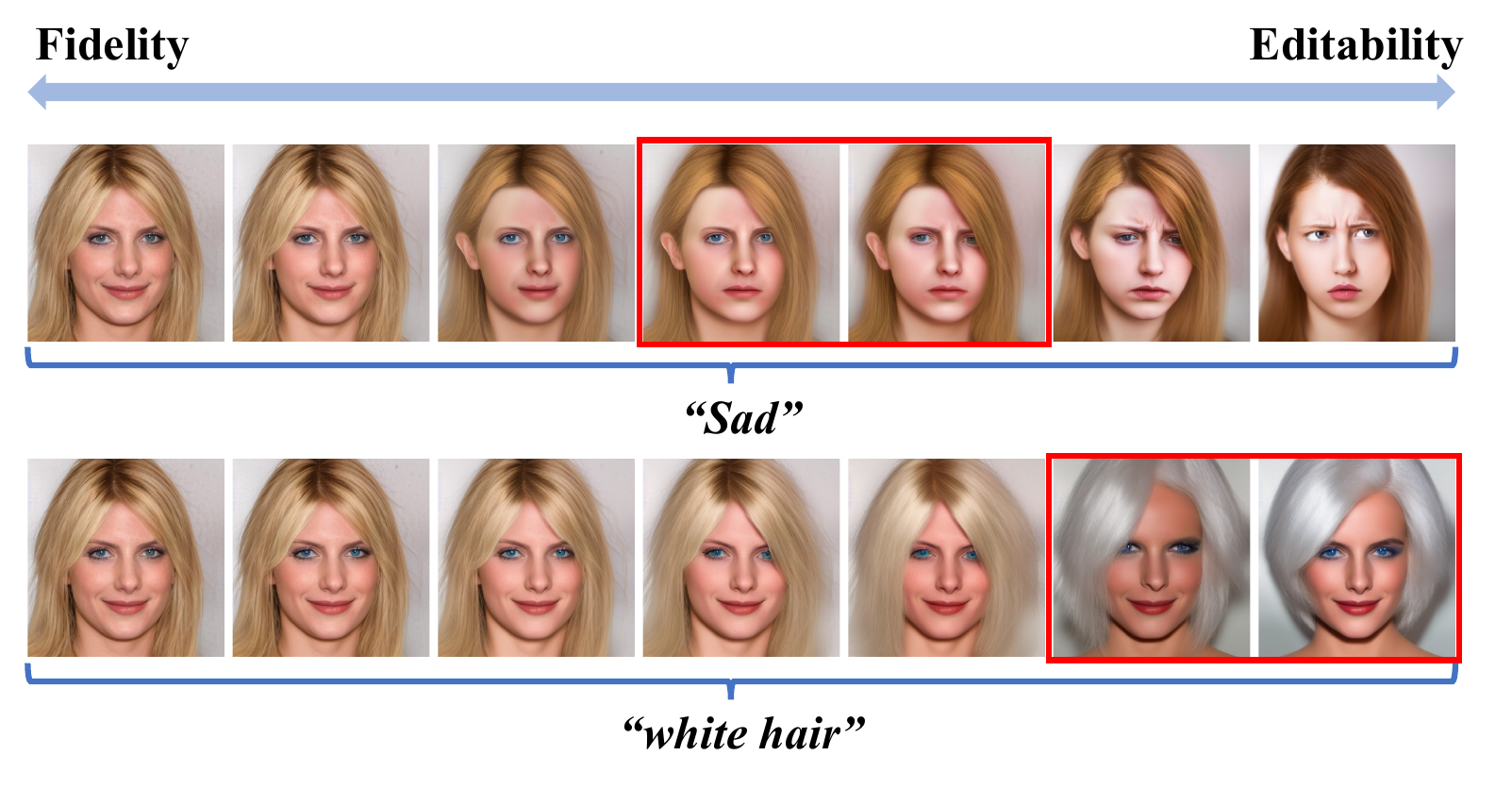}
    \caption{\textbf{Editability and fidelity trade-off of diffusion-based image manipulation.} The leftmost is the input image. For each manipulation, we add increasing noise levels from left to right and then denoise the image. Different semantics require different levels of noise to manipulate.}
    \label{fig:intro-tradeoff}
\end{figure}

A typical text-guidance manipulation pipeline is to invert the input image to a noisy latent and then denoise this latent with a given text prompt. The inversion process can be the simple noise adding \cite{meng2021sdedit} or the DDIM inversion \cite{su2022dual, song2020ddim}. 
There is an editability-fidelity tradeoff in such pipelines, as shown in Fig.~\ref{fig:intro-tradeoff}. 
Adding a lot of noise to the original image gives the diffusion model more freedom to manipulate the image, but it also makes it harder to retain original semantics when denoising back, and vice versa. 

More importantly, this trade-off can lead to unsuccessful manipulations, such as ``white hair" in Fig.~\ref{fig:intro-tradeoff}. This is because the diffusion model processes different semantics at different stages \cite{choi2022perception}. If our manipulation corresponds to the semantics in the early stages, we have to add more noise, hence losing much information from the original image.
To overcome the failure caused by the trade-off, existing methods attempt to incorporate more guidance information, such as using masks to limit the manipulation region \cite{avrahami2022blended, avrahami2022blendedlatent, lugmayr2022repaint}. This is useful for local image manipulation, but for some global structures, such as changing the pose of a human face, it still fails to solve the problem.

In this work, we take a different viewpoint to leverage a diffusion model for image manipulation.
Unlike existing methods that directly manipulate images progressively with diffusion models, we train an efficient image manipulator supervised by a pretrained diffusion model. 
Specifically, our model takes a feedforward model (e.g., latent space image manipulation model \cite{patashnik2021styleclip}) as the manipulator, and a text-guided diffusion model (e.g., latent diffusion \cite{rombach2022stablediffusion}) as the supervisor. During training, the manipulated image is diffused and fed to the diffusion model, and the diffusion model produces gradient supervision based on the text condition. In this sense, it is expected that the manipulator could mimic the promising generation capacity of the diffusion model.

To learn such a manipulator, using correct semantic guidance is also crucial, as shown in Fig. \ref{fig:intro-tradeoff}. 
Intuitively, the diffusion model at the timestep in the red box has a better ability to guide the image manipulator in changing the hair color. In contrast, other timesteps do not provide useful semantic guidance.
Therefore, we propose the hybrid quality score (HQS), an effective indicator that helps to select appropriate timesteps.
This indicator is built upon the entropy and $L_1$ norm of the gradient from the diffusion model and is shown to be highly correlated with the manipulation quality. As such, our model could learn with the most effective guidance. 
In addition to solving the trade-off problem of diffusion-based image manipulation, we have another bonus: our image manipulator requires only one forward for manipulation, and by learning in a specific domain, our image manipulator can manipulate images over the entire domain, which offsets the cost of training. Extensive experiments demonstrate the effectiveness of our methods.

Our contributions are summarized as follows:
\begin{itemize}
  \item We propose a novel image manipulation approach with a well-trained diffusion model to supervise another image manipulator. This avoids the trade-off problem of manipulating images with a diffusion process.
  \item We propose the hybrid quality score to detect semantic-related timesteps. Only during these timesteps can the diffusion model guide the image manipulator to perform accurate manipulations.
  \item Our experiments demonstrate our method's effectiveness and efficiency in both the qualitative and the quantitative aspects.
\end{itemize}

%% file: 02_related_v2.tex
\section{Related Work}
\label{sec:related}

Image manipulation \cite{isola2017image,choi2018stargan,chen2019homomorphic,chen2020domain,patashnik2021styleclip,kwon2022clipstyler,kocasari2022stylemc,avrahami2022blended,avrahami2022blendedlatent,choi2021ilvr,meng2021sdedit,zhao2022egsde,hertz2022prompt,kawar2022imagic,ruiz2022dreambooth} aims to modify an input image to a guiding direction. The direction can be a scribble, a mask, a reference image, etc.
Recently, text-driven image manipulations \cite{patashnik2021styleclip,kwon2022clipstyler,kocasari2022stylemc,avrahami2022blended,avrahami2022blendedlatent, hertz2022prompt,kawar2022imagic,ruiz2022dreambooth,gal2022stylenada} have become very popular because textual prompts are intuitive and easily accessible. 
These methods \cite{kwon2022clipstyler,kim2022diffusionclip,patashnik2021styleclip,kocasari2022stylemc,gal2022stylenada,avrahami2022blended} usually utilize a joint image-text semantic space to provide supervision. 

One of the famous image-text semantic spaces is CLIP \cite{radford2021CLIP}, a multi-modal space that contains extremely rich semantics as it is trained using its millions of text-image pairs. 
It has demonstrated significance on different tasks, such as latent space manipulation\cite{patashnik2021styleclip}, domain adaptation \cite{gal2022stylenada}, and style transfer\cite{kwon2022clipstyler}.
Another multi-modal model that has gained much attention is the text-image generative model. Several large-scale text-image models have advanced text-driven image generation, such as Imagen\cite{saharia2022imagen}, DALL-E2\cite{ramesh2022dalle2}, and latent diffusion\cite{rombach2022stablediffusion}.
The diffusion model contains the strong mode-capturing ability and the training stability \cite{ho2020ddpm,sohl2015deep}. 
Some scholars have begun to study how to utilize its powerful capabilities in image manipulation tasks. Most image manipulation methods with diffusion models aim to ``hijack'' the reverse diffusion process and introduce kinds of guidance and operation\cite{zhao2022egsde,meng2021sdedit,avrahami2022blended,avrahami2022blendedlatent,choi2021ilvr,lugmayr2022repaint,dhariwal2021guided, ho2022classifier}. 
For example, some methods \cite{dhariwal2021guided,zhao2022egsde, ho2022classifier,avrahami2022blended} update the intermediate result with the gradient from some conditional models. 
Some methods also use auxiliary information, such as a mask, to limit the region of the generation\cite{avrahami2022blended, lugmayr2022repaint}. 
Some method \cite{choi2021ilvr} directly replaces the low-frequency information of the intermediate result with that of the reference image and obtains an image with the same structural details.
Some method \cite{meng2021sdedit} conditions the output image with a stroke image by adding intermediate noises to the image and then denoising it.
Some method \cite{kim2022diffusionclip} applies domain adaptation to diffusion models using the CLIP model.
In summary, the diffusion process gradually perturbs the data distribution to gaussian noise distribution, while the reverse diffusion process progressively recovers data distribution from the noise. 

Meanwhile, another methods \cite{graikos2022diffusion,poole2022dreamfusion} uses the diffusion model as prior knowledge in many applications.
For example, some scholars \cite{graikos2022diffusion} mix it with conditional models, e.g., differentiable classifier, and generate specific classes of samples.
On the other hand, DreamFusion \cite{poole2022dreamfusion} utilizes differentiable image parameterization \cite{mordvintsev2018differentiable}  and defines this conditional model as a 3D rendering process \cite{mildenhall2020nerf} to generate 3D assets from the text. 
They can not involve its use in the inference stage\cite{poole2022dreamfusion}.
We also take the perspective to use the diffusion model as guidance to avoid the iterative process in the inference stage. 
Different from utilizing fixed conditional models \cite{graikos2022diffusion,poole2022dreamfusion}, we define this conditional model as an image-to-image translation model and explore a new scenario where we are optimizing this conditional model.
This model, after optimization, can be independent of the diffusion model, thus enabling more efficient image manipulation.

%% file: 03_background_v2.tex
\section{Preliminary}
\label{sec:preliminary}
\subsection{Diffusion Models}
\label{sec:diffusion-models}
Diffusion models are latent-variable generative models that define a Markov chain of diffusion steps to add random noise to data slowly and then learn to reverse the diffusion process to construct desired data samples from the noise \cite{sohl2015deep,ho2020ddpm}.  

Suppose the data distribution is $q(x_0)$ where the index $0$ denotes the original data. Given a training data sample $x_0\sim q(x_0)$, the forward diffusion process aims to produce a series of noisy latents $x_1,x_2,\cdots,x_T$ by the following Markovian process,
\begin{equation}
q(x_t\mid x_{t-1}) = \mathcal{N}(x_t;\sqrt{1-\beta_t}x_{t-1},\beta_t\mathbf{I}), \forall t\in [T],
\label{eq:forward}
\end{equation}
where $T$ is the step number of the diffusion process, $[T]=\{1,2,\cdots, T\}$ denotes the set of the index, $\beta_t\in(0,1)$ represents the variance in the diffusion process, $\mathbf{I}$ is the identity matrix with the same dimensions as the input data $x_0$, and $\mathcal{N}(x;\mu,\sigma)$ means the normal distribution with mean $\mu$ and covariance $\sigma$.

To generate a new data sample, diffusion models sample $x_T$ from the standard normal distribution and then gradually remove noise by the intractable reverse distribution $q(x_{t-1}\mid x_t)$. Diffusion models learn a neural network $p_{\theta}$ parameterized by $\theta$ to approximate the reverse distribution as follows,
\begin{equation}
p_{\theta}(x_{t-1}\mid x_{t}) = \mathcal{N}(x_{t-1};\mu_\theta(x_t, t),\Sigma_\theta(x_t, t)), 
\label{E3}
\end{equation}
where $\mu_\theta$ and $\Sigma_\theta$ are the trainable mean and covariance functions, respectively.

In \cite{ho2020ddpm}, $\Sigma_\theta$ is simply set as a fixed constant, and $\mu_\theta$ is reformulated as a function of noise as follows,
\begin{equation}
\mu_\theta(x_t,t)=\frac{1}{\sqrt{\alpha_t}}\left(x_t-\frac{\beta_t}{\sqrt{1-{\bar{\alpha}_t}}}\epsilon_\theta(x_t, t)\right), \\
\end{equation}
where $\epsilon_\theta$ is used to predict noise $\epsilon_t$ from $x_t$.

Finally, the diffusion model is trained with simplified evidence lower bound (ELBO) that ignores the weighting term for each timestep $t$ as follows,
\begin{equation}
\mathcal{L}_t(\theta)=\mathbb{E}_{x_0, t,\epsilon}\left[ \left\Vert \epsilon_t - \epsilon_{\theta}(x_t, t) \right\Vert^2_2\right].
\label{E5}
\end{equation}

\subsection{Diffusion Models as Prior}
\label{sec:diffusion-models-as-prior}
According to \cite{graikos2022diffusion}, diffusion models can also be used as off-the-shelf modules in some scenarios, where it $p(x)$ may serve as a prior for another conditional model $c(x, y)$, \ie we can deduce $p(x\mid y)$ given $p(x)$ and $c(x,y)$.
When $c(\mathbf{x},\mathbf{y})$ is a hard and non-differentiable conditional model, say a deterministic function $x=f(y)$. We optimize $y$ to minimize \footnote{Please refer supplementary for detailed derivation process.}
\begin{equation}
\begin{aligned}
\sum_t\mathbb{E}_{\epsilon\sim \mathcal{N}(0,I)}[\left \Vert \epsilon - \epsilon_{\theta}(\sqrt{\bar{\alpha}_t}f(y) + \sqrt{1-\bar{\alpha}_t}\epsilon,t) \right \Vert^2_2].
\end{aligned}
\label{E8}
\end{equation}
For example, $f()$ is a latent-variable model that takes latent $y$ as input and generates a sample $x$. 
We can regard this equation as sampling $y$ instead of directly sampling images using diffusion models, and inputting $y$ to this conditional model, we will get a sample from the diffusion model.
An example of a successful implementation of this idea is DreamFusion \cite{poole2022dreamfusion}, where $y$ represents the parameters of a 3D volume and $f()$ represents a volumetric renderer. This method can be used to generate 3D assets from text.
In practice, optimizing simultaneously for all $t$ makes it difficult to guide the sample toward a mode. Thus existing methods eighter anneal $t$ from high to low values \cite{graikos2022diffusion}, or random select $t$ \cite{poole2022dreamfusion}.
So the actual optimization process is slightly changed to
\begin{equation}
\begin{aligned}
\mathbb{E}_{\epsilon,t}[\left \Vert \epsilon - \epsilon_{\theta}(\sqrt{\bar{\alpha}_t}f(y) + \sqrt{1-\bar{\alpha}_t}\epsilon,t) \right \Vert^2_2].
\end{aligned}
\label{E9}
\end{equation}

%% file: 04_method_v7.tex
\section{Method}
\label{sec:method}
\begin{figure}[t]
\centering
\includegraphics[width=1.0\linewidth]{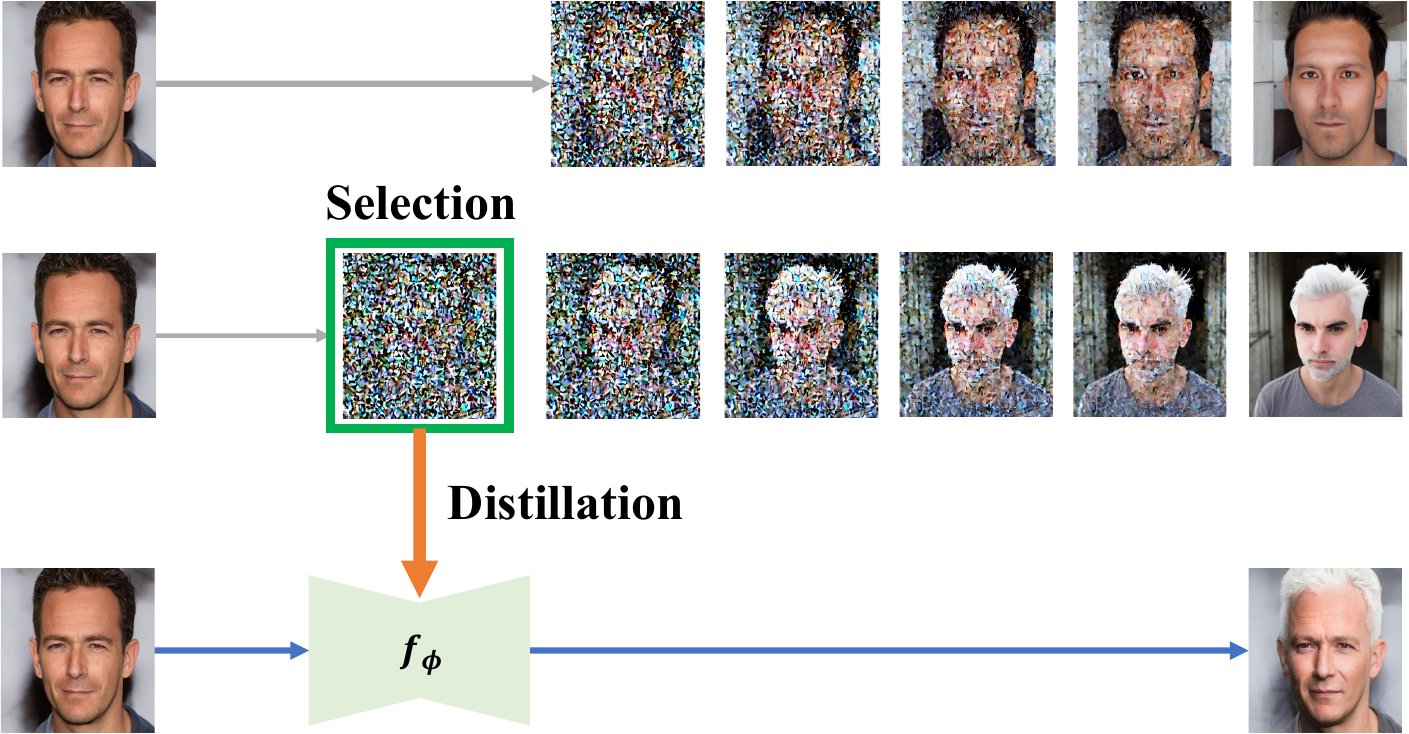}
\vspace{-0.2in}
\caption{\textbf{Core concept of SDD.} Our method involves two steps: 1) selecting the semantically-related timestep and 2) distilling the appropriate semantic knowledge into an image manipulator, $f_{\phi}$.}
\label{fig:method-SDD}
\vspace{-0.1in}
\end{figure}

In this section, we introduce Selective Diffusion Distillation. The core concept is shown in Fig.~\ref{fig:method-SDD}. First, we introduce how to distill knowledge from a diffusion model into an image manipulation model in Sec.~\ref{sec:distillation}. Then, we introduce how to select the appropriate timestep in Sec.~\ref{sec:selection}.

\subsection{Distillation: Learning image manipulator with Diffusion Models}
\label{sec:distillation}

\begin{figure*}[htbp]
    \centering
    \includegraphics[width=17cm]{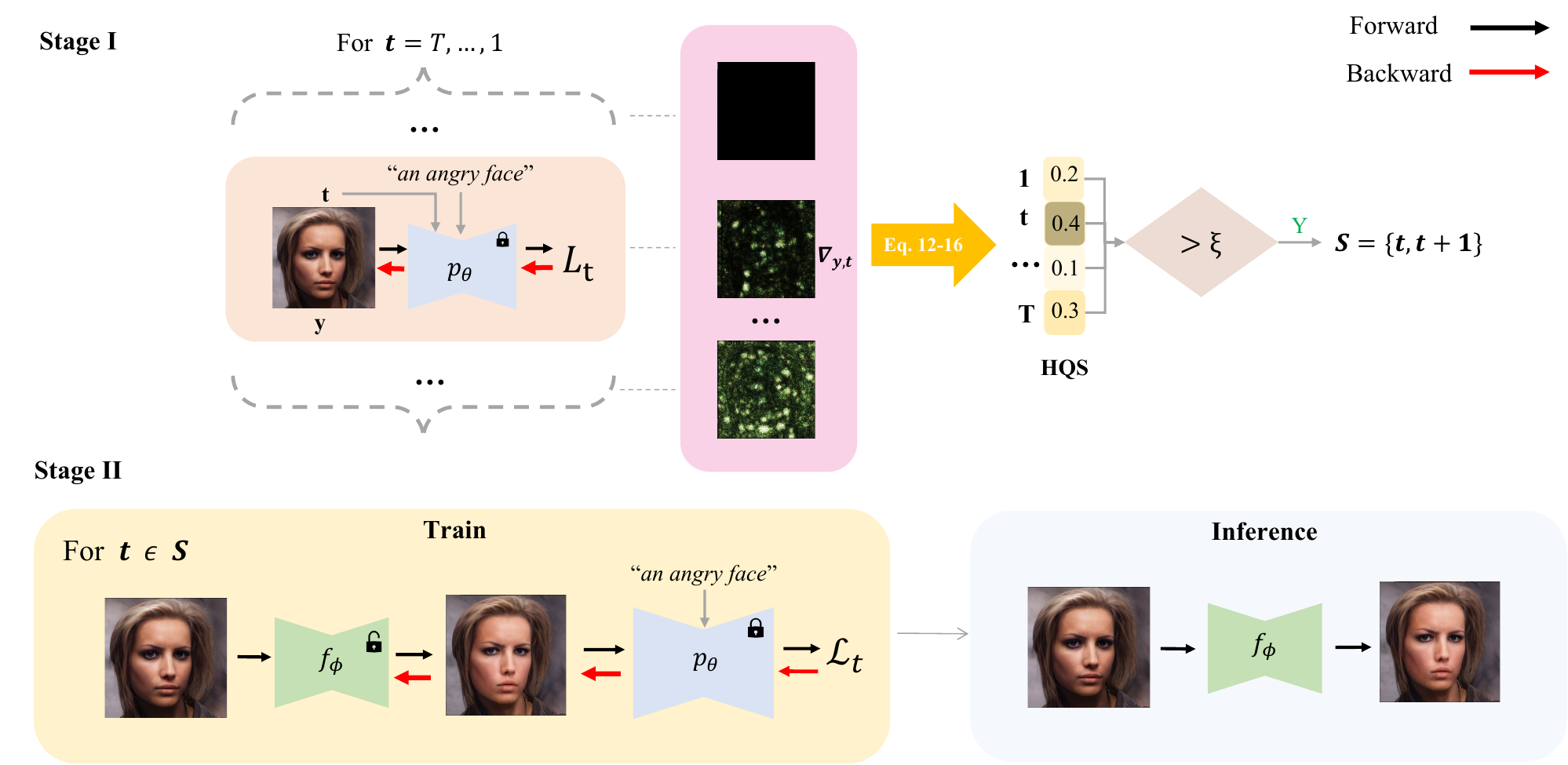}
    \caption{ \textbf{General framework of Selective Diffusion Distillation (SDD)}. \textbf{Top: selection stage}. We build an HQS indicator to select appropriate timesteps. \textbf{Bottom: distillation stage}. We use the selected timesteps and the pretrain diffusion model $p_\theta$ to train the image manipulator $f_\phi$.}
    \label{fig:method-framework}
\end{figure*}

We aim to take the diffusion model as a source of semantic guidance to train a lightweight image manipulator.
We first define our problem according to the definition in Sec. \ref{sec:diffusion-models-as-prior}. 
Training the image manipulator with diffusion prior can also be categorized as an optimization problem with a hard constraint. The hard constraint $f(y)$ becomes the image manipulator in this situation. The difference between our formulation and the previous formulation is the optimized target. We use equation \eqref{E10} to optimize parameters $\phi$ of our image manipulator $f_\phi$ as shown in Fig. \ref{fig:method-framework}. 

\begin{equation}
\begin{aligned}
\min_{\phi}\mathbb{E}_{\epsilon,t}[\left \Vert \epsilon - \epsilon_{\theta}(\sqrt{\bar{\alpha}_t}f_\phi(y) + \sqrt{1-\bar{\alpha}_t}\epsilon,t) \right \Vert^2_2].
\end{aligned}
\label{E10}
\end{equation}

When optimizing this equation, we follow the approach of skipping the U-Net Jacobian in \cite{poole2022dreamfusion}. The gradient after skipping is equivalent to the noise predicted at the current timestep (based on the diffusion model) minus a random noise. 
According to the relationship between diffusion models and score-based models\cite{song2019score1,song2020score2}, this predicted noise from the diffusion model contains the direction from the current distribution to the target distribution.
Therefore, we can consider optimizing this equation as guiding our image manipulator by the diffusion model to output results in the target distribution.

This paradigm benefits us from two aspects.
Firstly, no matter what semantics we manipulate, we can always ensure fidelity.
For general diffusion-based image manipulation, the success of manipulation and the fidelity of the manipulated image are both determined by the noise level, but they conflict with each other. However, for our image manipulator, this conflict does not exist. It is natural to add or tweak regularization terms in the training of our image manipulator so that we can ensure fidelity under different manipulations.
Secondly, our method improves inference efficiency. After optimizing the image manipulator, we can perform fast image manipulation through only one forward pass without requiring the diffusion model. Not to mention the manipulator network is lighter than the U-net in the diffusion model.
Moreover, our manipulator also demonstrates scalability. Our manipulator finds common knowledge when translating images in the same domain to a specific semantic direction. Once the manipulator is trained, it is easy to reuse this network for manipulating other images without any retraining.
Compared to the efficiency improvement, our extra training cost is little. As described in Sec.~\ref{sec:experiments}, when training the manipulator, we only optimize a 4-layer MLP mapper model. This significantly reduces our training costs. 
When manipulating more images, our method shows faster speed even if we include our extra training time. Detailed quantitative discussion can be found in Sec.~\ref{sec:exp-comparison}.

\subsection{Selection: selecting timestep with the Hybrid Quality Score (HQS)}
\label{sec:selection}
Assuming fidelity is ensured, the key is to obtain correct semantic guidance from the diffusion model. To achieve this, we need to identify the timestep at which the diffusion model most effectively guides the image manipulator to produce a successful result.
We first analyze the gradients provided by the diffusion model. We input the image $y$ into the diffusion model conditioned on textual description $\gamma$ at every timestep $t\in\{T,\cdots,1\}$, and then compute the gradient on the input image using diffusion training loss:
\begin{eqnarray}
d_t(y,\gamma) = \nabla_{y}\left \Vert \epsilon - \epsilon_{\theta}(\sqrt{\bar{\alpha}_t}y + \sqrt{1-\bar{\alpha}_t}\epsilon,t,\gamma) \right \Vert^2_2,
\label{grad}
\end{eqnarray}
where $d_t(y,\gamma) \in \mathbb{R}^{1\times H\cdot W\cdot C}$ represents the direction that the input image $y$ should move to, given the target distribution $\gamma$ at the timestep $t$.
If a timestep $t$ is more important, its direction $d$ should have higher quality than other timesteps.

To measure its quality, we have empirically observed that treating this gradient as a confidence map and computing its entropy gives us a good metric.
We first convert $d_t$ into a confidence score map $p_t$ through the softmax operation.
Each score of the confidence map represents the degree of necessity of the corresponding gradient when modifying the image. 
Next, for the whole confidence map, we calculated its entropy:
\begin{equation}
H_t = -\sum_{i}p^i_t \log p^i_t,
\label{E13}
\end{equation}
where $p_t^i$ is the i-th element of $p_t$.
The intuition is that the lower the entropy, the more the $d_t(y,\gamma)$ contains the necessary gradient, so the $t$ is more important.
Then, we found this metric susceptible to extreme cases, such as when only a very small part of the gradient map has value. 
This will result in a very high value of $H_t$, but the gradient contributes little to the image.
Therefore, we introduce the $L_1$ norm of the gradient $d_t(y,\gamma)$ to avoid this situation:
\begin{equation}
N_t = \sum_{i} |d_t^i|,
\label{E14}
\end{equation}
where $d_t^i$ is the i-th element of $d_t$.
This metric ensures the magnitude of the overall information of the gradient, thereby avoiding the misjudgment of the entropy metric caused by the local large value.
To combine these two metrics, we firstly transform $H = [H_1,\cdots, H_T]$ to $\bar{H} = [\bar{H}_1,\cdots,\bar{H}_T]$ by using min-max normalization as follows:
\begin{equation}
\bar{H}_t = \frac{H_t - \mathrm{min}(H)}{\mathrm{max}(H) - \mathrm{min}(H)},
\label{E15}
\end{equation}
and we also do the same to $N$ to get $\bar{N}$.
\begin{figure}[tbp]
    \centering
    \includegraphics[width=8cm]{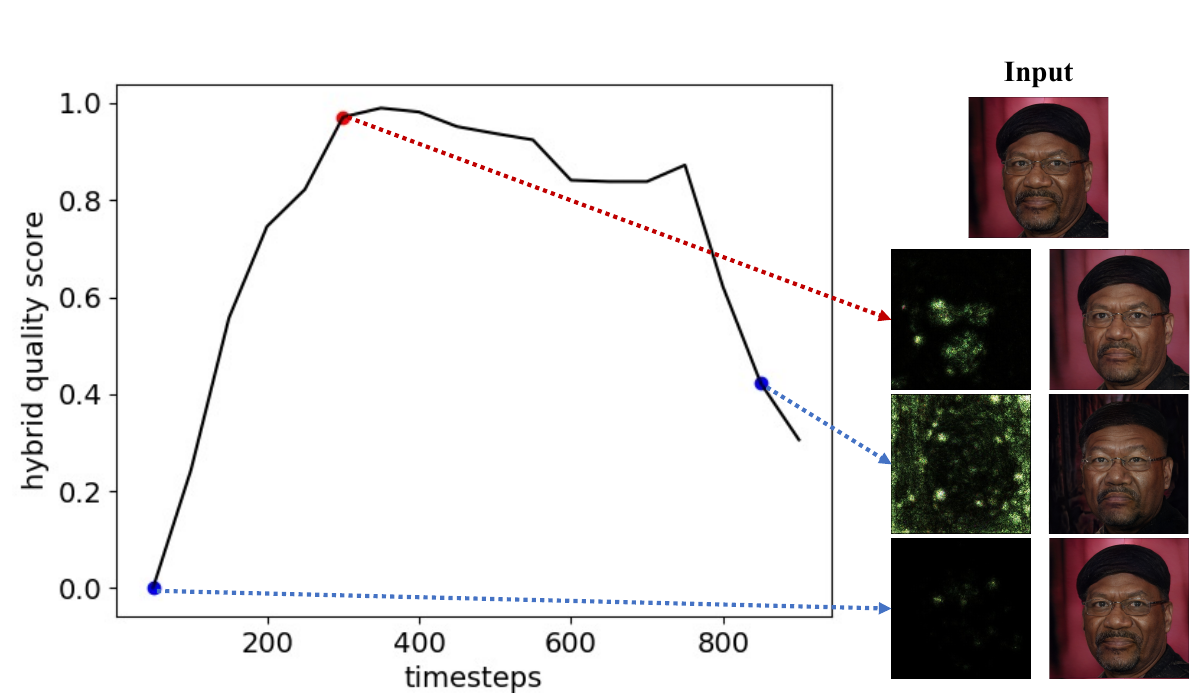}
    \caption{\textbf{Effect of our Hybrid Quality Score (HQS)}. \textbf{Left}: The curve of HQS at different timesteps when the prompt is ``angry". \textbf{Right}: Gradient visualization of the corresponding timestep and editing results of training our image manipulator using only this timestep.}
    \label{HQS}
\end{figure}
Significantly, given manipulation target $\gamma$, we consider timestep $t$ with lower $\bar{H}_t$ and higher $\bar{N}_t$ to be optimal, so we define a metric called Hybrid Quality Score ($\mathrm{HQS}$) :
\begin{equation}
\mathrm{HQS}(\gamma) = \mathbb{E}_{y}[\bar{N} - \bar{H}],
\label{eq:HQS}
\end{equation}
Where $\mathrm{HQS} \in \mathbb{R}^{1 \times T}$.
As shown in Fig. \ref{HQS}, training with the timestep with the higher hybrid quality score gives us rich semantic gradients\footnote{we clip and normalize the gradient value for better illustration.} and better editing results. 

Based on this, we propose a timestep selection strategy:
Given text prompt $\gamma$, we compute the $\mathrm{HQS}(\gamma)$ at each timestep $t$. Then we build a set of $t$ with a larger $\mathrm{HQS}$ value.
Next, when optimizing the image manipulator with Eq.~\eqref{E10}, we sample timesteps $t$ from the set.
To decide the number of timesteps in this set, we introduce a hyperparameter, $\xi$, which controls the tolerance for uncorrelated $t$. While selecting $t$ with the maximum $\mathrm{HQS}$ is generally effective, relaxing the $\mathrm{HQS}$ requirement can also improve editing. By using a smaller $\xi$, only the most relevant features will change, making the semantic modification relatively simple but requiring fewer optimization costs. On the other hand, using a larger $\xi$ makes the editing more comprehensive but increases the risk of introducing uncorrelated $t$.

In conclusion, training our image manipulator $f_\phi$ with this strategy is shown in Alg. \ref{alg:cap}. 

\begin{algorithm}
\caption{Selective Diffusion Distillation}\label{alg:cap}
\begin{algorithmic}[1]
\State {\bfseries Input:} text prompts $\gamma$, image data $q(y)$, threshold $\xi$, 
 pretrained diffusion model $\epsilon_{\theta}$
\State Compute $\mathrm{HQS}(\gamma)$ by Eq. \eqref{eq:HQS}
\State $S=\{t \mid \mathrm{HQS}_t > \xi \}$
\State Randomly initialize our image manipulator $f_\phi$
\Repeat
    \State $y \sim q(y)$, $t \in S$, $\epsilon \sim \mathcal{N}(\mathbf{0},\mathbf{I})$
    \State Take gradient descent step on 
    \State $\qquad \nabla_{\phi}\left \Vert \epsilon - \epsilon_{\theta}(\sqrt{\bar{\alpha}_t}f_\phi(y) + \sqrt{1-\bar{\alpha}_t}\epsilon,t) \right \Vert^2_2$
\Until converged
\end{algorithmic}
\end{algorithm}

Using this strategy leads to a set $S$ of a smaller size. This selected $S$ contains fewer ineffective timesteps than the whole timesteps set. Training the image manipulator with selected $S$ leads to correct semantic manipulation.
We prove the effectiveness in the Sec. \ref{sec:ablation}. The above analysis is agnostic to the concrete form of image manipulators so that it can be extended to any image manipulation framework. 

%% file: 05_experiment_v1.tex
\section{Experiments}
\label{sec:experiments}

In this section, we first introduce the implementation details of our Selective Diffusion Distillation. 
Then, we present the manipulating ability of SDD by showing its application in different domains.
Afterward, we compare our method to other image manipulation methods regarding quality and efficiency. 
At last, we conduct the ablation study of HQS-based step selection to demonstrate its effectiveness.

\subsection{Implementation details}
\label{sec:implementation}
The discussion in Sec.~\ref{sec:distillation} shows that our framework is independent of the form of the image manipulator. Therefore, our method can theoretically be applied to any image manipulation method. 

We select StyleGAN\cite{karras2019style} as our image manipulator backbone because of its exceptional capabilities in image manipulation\cite{gal2022stylenada, wu2021stylespace, patashnik2021styleclip, kocasari2022stylemc, richardson2021encoding}. In our framework, both the diffusion model and the StyleGAN contain large parameters. Directly optimizing the StyleGAN by our framework could produce unacceptable computational costs. Therefore, we follow the configuration of \cite{patashnik2021styleclip} to reduce the computational cost. 
Under this configuration, we only optimize a tiny MLP, called latent mapper, to achieve numerous editing tasks. 

For the text-to-image diffusion model, we employ the popular latent diffusion models \cite{rombach2022stablediffusion}.

Overall, our image manipulator consists of three components: a StyleGAN encoder, a latent mapper, and a StyleGAN generator. The StyleGAN encoder and generator are pre-trained and remain fixed during the optimization. only the latent mapper is trained in the distillation.

For regularization, we introduce L2 loss and face identity loss\cite{deng2019arcface} as suggested by \cite{patashnik2021styleclip}. We also use gradient clipping in some scenarios for training stability. The whole training procedure follows the Alg.~\ref{alg:cap}.

\begin{figure*}[t]
\centering
\includegraphics[width=1.0\linewidth]{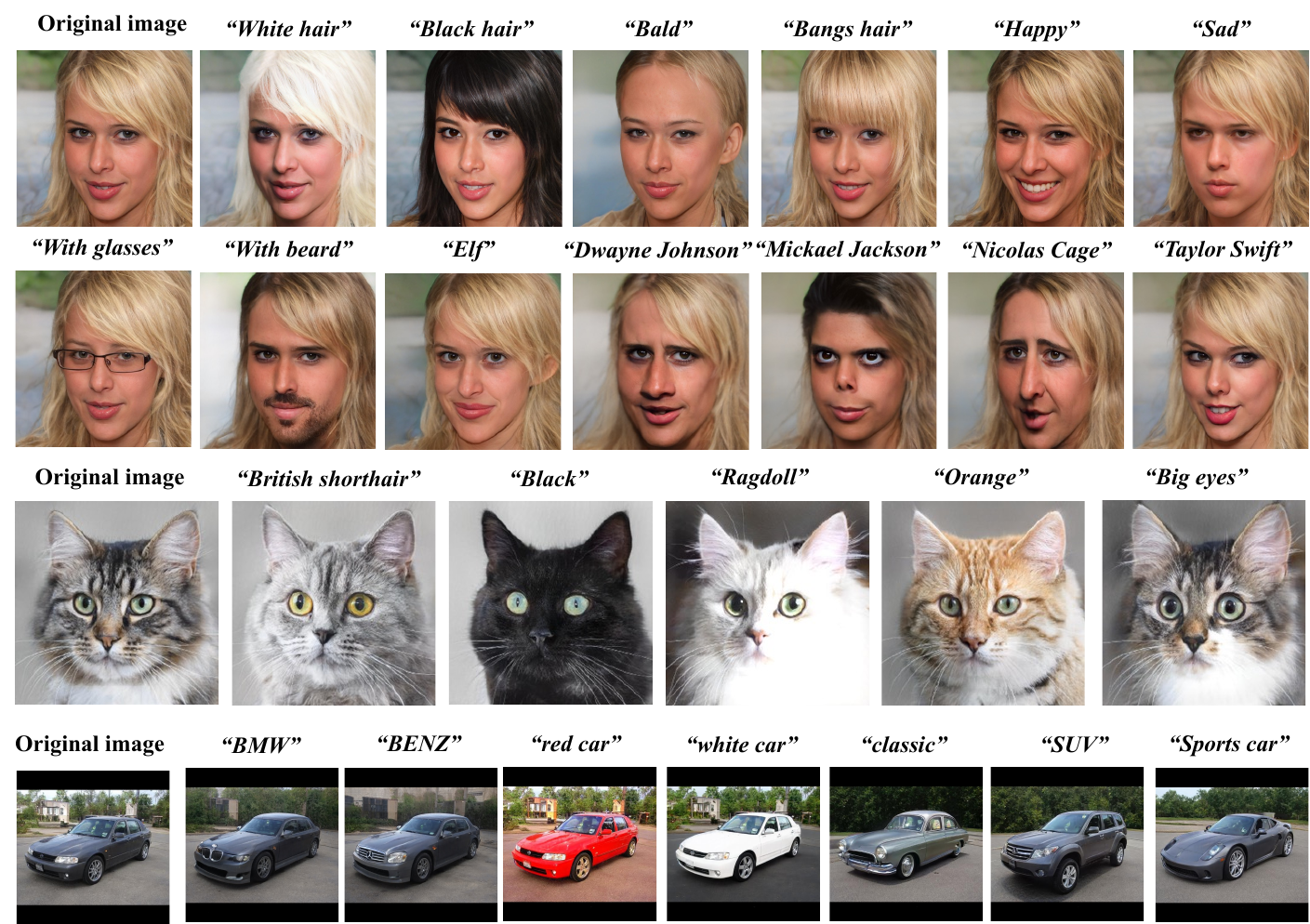}
\vspace{-0.2in}
\caption{The manipulation results of SDD of various domains (CelebA-HQ\cite{liu2015faceattributes}, AFHQ-cat\cite{choi2020stargan}, LSUN-car\cite{yu2015lsun}) and various attributes. We keep the image fidelity and make it coherent with the text.}
\label{exp-showcase}
\vspace{-0.1in}
\end{figure*}
\subsection{Applications}
\label{sec:application}
Our SDD shows its strong ability in image manipulation. It successfully edits images of various domains and various attributes while preserving high image quality. Fig.~\ref{exp-showcase} demonstrates our manipulated images. For human faces, we could conduct the hair color, hairstyle, and facial expression transfer, as well as the attributes addition and celebrities conversion. For cats and cars, we could change their types, colors, and some specific details. More manipulation results are shown in our supplementary materials.
 
\subsection{Comparison and Evaluation}
\label{sec:exp-comparison}
In this section, we compare SDD with other image manipulation methods. The empirical results demonstrate our benefits.
To measure the result quantitatively, we adopt two metrics, the Fréchet inception distance(FID) \cite{Seitzer2020FID} and CLIP similarity\cite{radford2021learning}, separately. FID is a metric used to measure the similarity between the distribution of real images and generated images. We use it to measure the similarity between the manipulated and original images. The CLIP similarity measures whether the semantic change of a manipulated image aligns with the text description. A higher directional CLIP similarity indicates a better manipulation result.

\begin{figure}[t]
\centering
\includegraphics[width=1.0\linewidth]{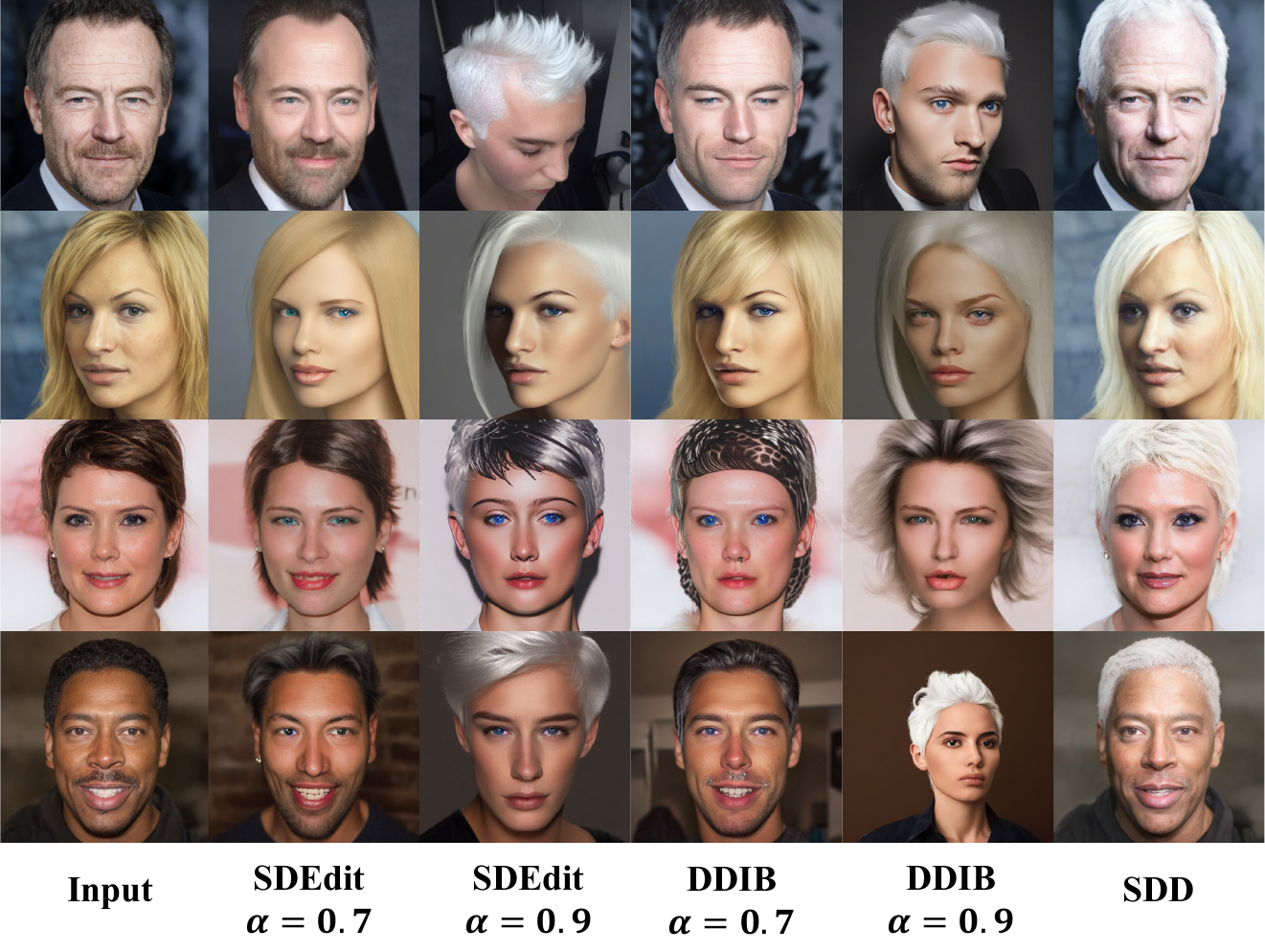}
\vspace{-0.1in}
\caption{\textbf{The visual comparison between SDD and diffusion-based image manipulations.} $\alpha$ ranges from zero to one and represents the noise level. SDEdit and DDIB fail to manipulate the semantics at a noise level of 0.7, while SDD successfully manipulates the semantics with less distortion.}
\label{fig:exp-comparison-diffusion}
\vspace{-0.1in}
\end{figure}

\paragraph{Compared to diffusion-based methods}
We first compare our SDD with other typical diffusion-based image manipulation methods, including SDEdit~\cite{meng2021sdedit}, DDIB~\cite{su2022dual}, and DiffAE~\cite{preechakul2022diffusion}. 
The qualitative is shown in Fig.~\ref{fig:exp-comparison-diffusion}. SDD achieves semantic manipulation in all cases while preserving most of the other information from the input images compared to the baselines.

The quantitative result is shown in Table~\ref{table:exp-diff-comparison}.
SDD demonstrates the highest CLIP similarity and maintains the lowest FID, which is consistent with the quantitative comparison. This excellent result indicates that SDD successfully avoids the trade-off problem in diffusion-based manipulation. Moreover, considering that the StyleGAN of our image manipulator is pre-trained with data of the target domain, we also finetune the best of the baselines on this data for a fair comparison. Line 2 of Table~\ref{table:exp-diff-comparison} shows that despite the extra data, SDD still outperforms it.




In addition, we also compare SDD with these baselines in terms of computation costs. Given the task of manipulating $\mathbf{m}$ images for each of $\mathbf{n}$ prompts, we compute the total time cost of our method and diffusion-based method to compare their efficiency.
For diffusion-based methods, we consider the input image inversion time and the iterative denoising time as their total inference time. 
$\tau_{\text{Diff,infer}}$ denotes the time cost of each manipulation in diffusion-based methods. 
\begin{equation}
\begin{aligned}
\tau_{\text{Diff}} = m \times n \times \tau_{\text{Diff,infer}}
\end{aligned}
\label{E15}
\end{equation}

For SDD, we need to train individual image manipulators for each of the prompts, and when inference, only the image manipulator is needed. We add the training time of image manipulators and the inference time of manipulation together and treat it as the total time of SDD.
$\tau_{\text{SDD,train}}$ denotes the required training time for SDD's image manipulator to converge.
$\tau_{\text{SDD,infer}}$ denotes the time cost of inference with the image manipulator. 
\begin{equation}
\begin{aligned}
\tau_{\text{SDD}} = n \times \tau_{\text{SDD,train}} + n \times m \times \tau_{\text{SDD,infer}}
\end{aligned}
\label{E16}
\end{equation}
Accorind the Eq.~\ref{E15} and Eq.~\ref{E16}, all methods have the same complexity of $O(mn)$. However, in our method, the coefficient of $mn$ is $\tau_{\text{SDD,infer}}$, which is a much smaller value compared to $\tau_{\text{Diff,infer}}$.

We further deduce that when $m$ satisfied Eq. \eqref{E17}, our methods will achieve better efficiency. This efficiency improvement will be exaggerated, especially when $m$ becomes larger.
\begin{equation}
\begin{aligned}
m > \frac{\tau_{\text{SDD,train}} }{\tau_{\text{Diff,infer}} - \tau_{\text{SDD,infer}}}
\end{aligned}
\label{E17}
\end{equation}
We compared the overall time cost between our method and diffusion-based methods in Table~\ref{table:exp-diff-comparison}. All diffusion models use DDIM\cite{song2020ddim} acceleration with 50 inference steps. We set m=100, and n=10 in the experiment. The comparison of computational costs is shown in column 3 of Table~\ref{table:exp-diff-comparison}.
\begin{table}[hb]
\centering
\resizebox{0.9\linewidth}{!}{
\begin{tabular}{cccc}
\hline
        & FID            & CLIP Similarity  & Total time      \\ \hline
SDEdit  & 32.126         & 0.2189                    & 2215         \\
SDEdit* & 16.761         & 0.2133                   & 2215         \\
DDIB    & 87.737               & 0.2268              & 3502         \\
DiffAE  & 41.896         & 0.2136                    & 5658            \\ \hline
SDD     & \textbf{6.066} & \textbf{0.2337}  & \textbf{148.67} \\ \hline
\end{tabular}}
\caption{The quantitative comparison between our method and diffusion-based image manipulations. * means we fine-tune the diffusion model on the target dataset.}
\label{table:exp-diff-comparison}
\vspace{-0.2in}
\end{table}

\begin{figure}[t]
\centering
\includegraphics[width=0.8\linewidth]{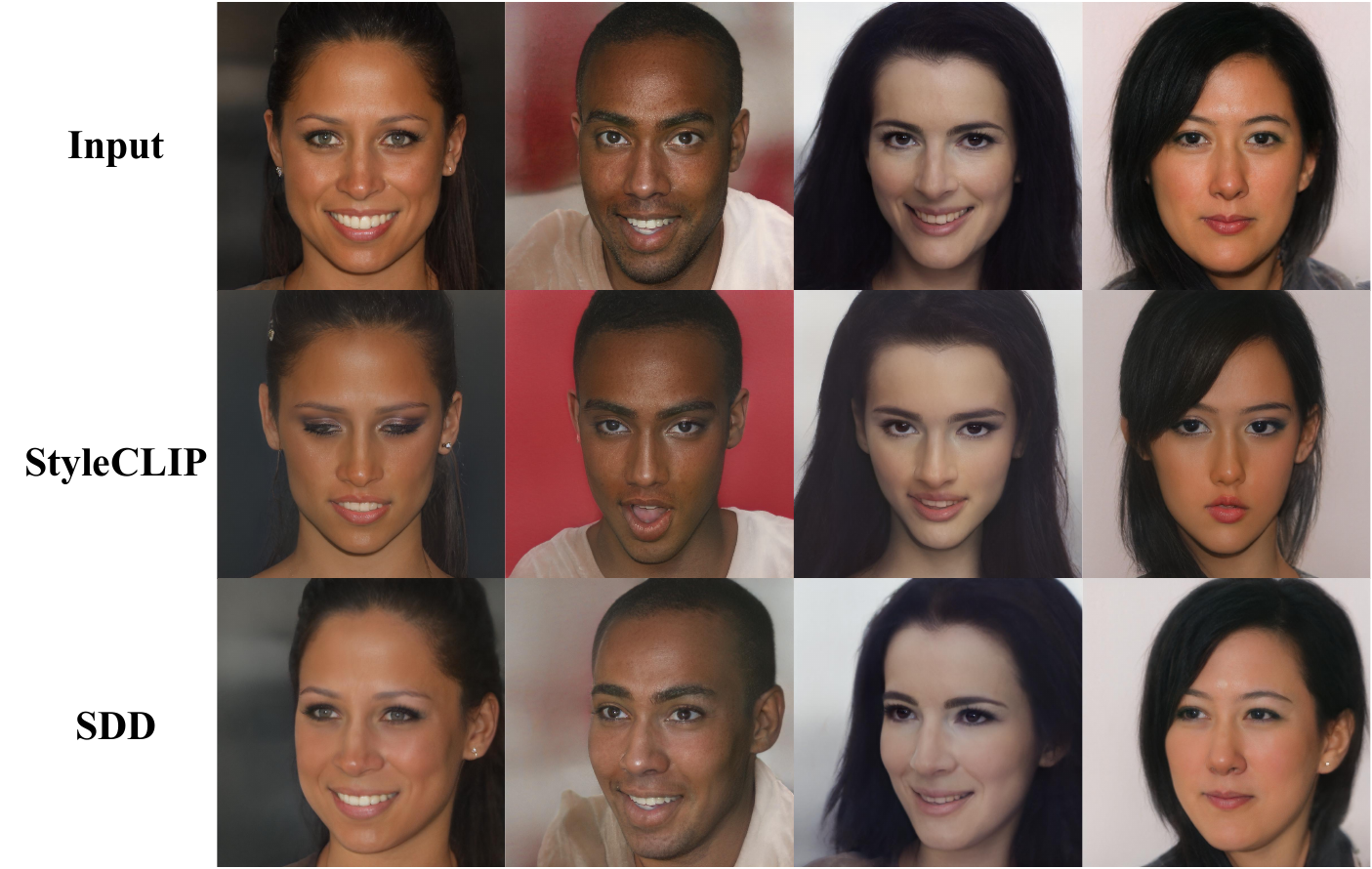}
\vspace{-0.1in}
\caption{\textbf{Comparison of SDD and StyleCLIP in pose manipulation}. Text prompt: ``side view''. The results demonstrate that the diffusion model provides superior semantic guidance, enabling a broader range of manipulations.}
\label{fig:exp-comparison-styleclip}
\vspace{-0.2in}
\end{figure}

\paragraph{Compared to StyleCLIP}
Our method shares some similarities to the StyleCLIP in the aspect of the manipulator. Here, we discuss the major difference between SDD and StyleCLIP and empirically compare them.
The diffusion model has a significant advantage over CLIP in that the gradient it provides shares the same spatial size as images. Compared to CLIP, these gradients from diffusion models contain structural information, making our SDD capable of position manipulations. In contrast, CLIP guidance is insensitive to 3D positional information, so it does not support such manipulation. Other studies \cite{Jain_2021_ICCV} also notice this insensitivity of CLIP, and they provide further explanation for that. 
Concretely, our image manipulator can change the pose of a human's face, but CLIP-based methods fails, as shown in Fig.~\ref{fig:exp-comparison-styleclip}. 

We also quantitatively compare manipulation tasks that can be accomplished by both methods and demonstrate that our SDD provides better guidance for training the image manipulator, as shown in Table~\ref{table:comparison}. Note that directly replacing CLIP with a diffusion model and not using our timestep selection strategy will not yield such results.
\begin{table}[htbp]
\begin{center}
\begin{tabular}{cccc}
\hline
          & Domains & FID    & CLIP Similarity \\ \hline
          & Face    & 16.542 & 0.2250          \\
StyleCLIP & Car     & 52.356 & 0.2652          \\
          & Cat     & 42.737 & 0.2582          \\ \hline
          & Face    & \textbf{6.066}  & \textbf{0.2337}          \\
SDD       & Car     & \textbf{52.356} & 0.2621          \\
          & Cat     & \textbf{39.354} & \textbf{0.2948}          \\ \hline
\end{tabular}
\caption{Quantitative comparison between SDD and StyleCLIP.}
\label{table:comparison}
\end{center}
\end{table}

\subsection{Ablation Study}
\label{sec:ablation}
In this section, we conduct an ablation study of the timestep selection strategy for optimizing the image manipulator. We build four configurations: For random strategy, we randomly sample $t$ from all timesteps; For small threshold strategy, we use HQS with a small $\xi$ to obtain $S$ with a large number of $t$; For large threshold strategy, we use HQS with a large $\xi$ to obtain $S$ with a small number of $t$; For largest HQS strategy, we only sample $t$ with the largest HQS value.
We keep other configurations the same and compare them qualitatively and quantitatively. 

\begin{figure}[tbp]
\centering
\includegraphics[width=8cm]{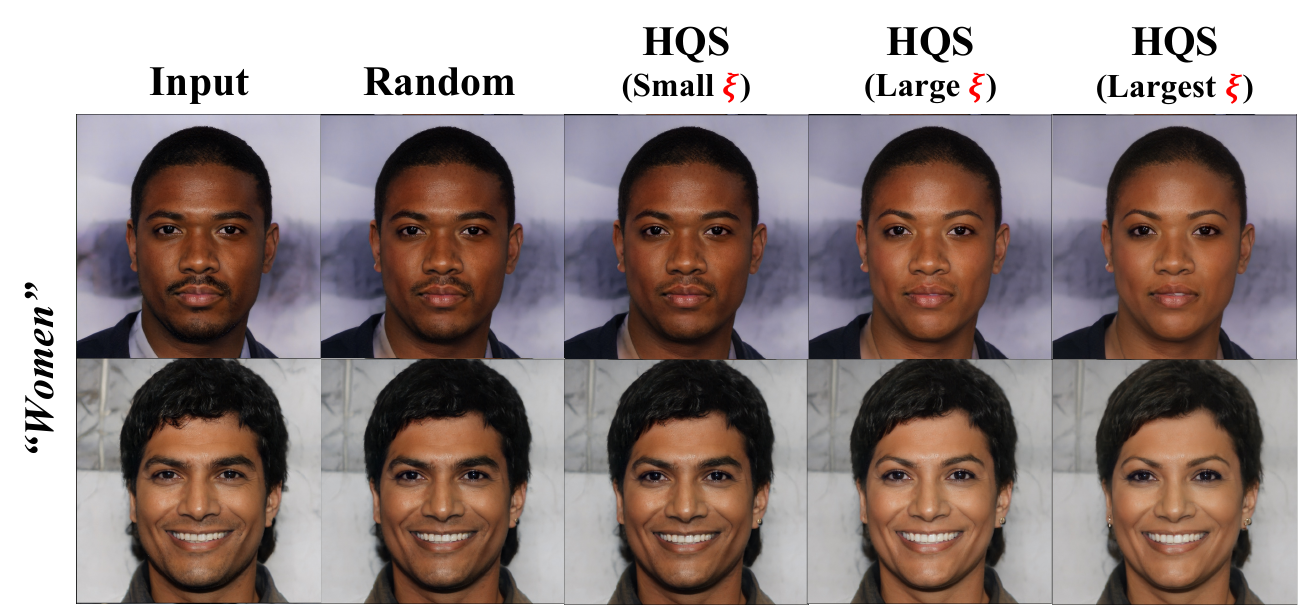}
\caption{\textbf{Visual result of Ablation Study}. The average $\mathrm{HQS}$ score of timesteps used for training increases from left to right and so does the accuracy of manipulation.}
\label{fig:exp-ablation-visual}
\vspace{-0.1in}
\end{figure}

The qualitative result in Fig.~\ref{fig:exp-ablation-visual} shows that the proposed $\mathrm{HQS}$-based step selection significantly overperforms other baseline methods. Random strategy always seems to have little modification to the original image. We attribute this to its redundant timestep selection. The result of the largest $\mathrm{HQS}$ strategy in the most desirable and intensive modification, proving that our $\mathrm{HQS}$ helps us find the most useful step. 
Furthermore, by combining the results from the small threshold, large threshold, and largest $\mathrm{HQS}$, we can observe that under the same number of training iterations, the average $\mathrm{HQS}$ score of the $t$ they sampled increased in order, leading to a sequential improvement of the manipulation effect. 
Therefore, it is proved that $\mathrm{HQS}$ can select the $t$ with the maximum contribution to the semantic manipulation. The quantitative result in Table~\ref{table:exp-ablation} also demonstrates that the largest HQS strategy performs the best. The increase in FID is caused by the manipulation effect, as shown in Fig.~\ref{fig:exp-ablation-visual}. Meanwhile, we still preserve image quality, as evidenced by our very low FID.

\begin{table}[h]
\centering
\begin{tabularx}{0.8\linewidth}{cccc}
\hline
                &FID  & CLIP Similarity \\ \hline
Random          &3.288 & 0.2146                \\
Small threshold &3.819& 0.2155                \\
Large threshold &5.927 & 0.2168                \\ 
Largest HQS     &9.154 & \textbf{0.2190}                \\ \hline
\end{tabularx}

\caption{The quantitative ablation for HQS-based timestep selection strategy. The FID increase because the manipulation effect increase, as shown in Fig.~\ref{fig:exp-ablation-visual}}
\label{table:exp-ablation}
\vspace{-0.2in}
\end{table}

%% file: 10_conclusion.tex
\section{Conclusion}
\label{sec:conclusion}
In this paper, we present a novel image manipulation method called Selective Distillation Diffusion (SDD). This paradigm avoids the Editability \& Fidelity trade-off by distilling the diffusion model's knowledge to a lightweight image manipulator. To distillate correct semantic information, we carefully design the Hybrid Quality Score (HQS) to help us select the helpful timesteps. We evaluate our method SDD on a variety of image manipulation tasks and achieve 
promising results.

%% file: 11_acknowledgement.tex
\section{Acknowledgement} This work is supported by National Natural Science Foundation of China (No. 62206068).

%% file: 12_appendix.tex
\appendix
\label{sec:appendix}

\section{Implementation Detail}
The image resolution for the human face dataset is set to $1024 \times 1024$.
The image resolution for the cat face and car dataset is set to $512 \times 512$.
The hyperparameters like learning rate, weight decay, loss weight, noise type, iterations, and mapper levels are adjusted individually for each prompt based on our experimental heuristics. 
More details can be found in our code. It is attached with the supplementary file.
All experiments are conducted on an NVIDIA RTX3090, with 24GB memory. The architecture of the image manipulator is the same as the latent mapper of \cite{patashnik2021styleclip}.

\section{Diffusion model as a prior}
As mentioned in Sec.~\ref{sec:diffusion-models-as-prior}, Diffusion models can also be used as off-the-shelf modules in some scenarios, where one model may be a prior for another conditional model. A typical example is a diffusion model $p(x)$ trained on MNIST digits and an off-the-shelf classifier $c(x, y)$ where $y$ is the class label. 
Then we can use the diffusion model to generate data of a specific class for the classifier.
In theory, this means that we want to deduce $p(x\mid y)$ given $p(x)$ and $c(x,y)$.
One solution is to introduce an approximate variational posterior $q(x)$ to approximate the posterior distribution $p(x|y)$, and minimize:
\begin{equation}
\small
\begin{aligned}
F = -\mathbb{E}_{q(x)}[\log{p(x)}-\log{q(x)}]-\mathbb{E}_{q(x)}[\log{c(x,y)}]
\end{aligned}
\label{E6}
\end{equation}
When extending this formula to the scenario of diffusion model with latent variable $x_1,...,x_T$, we can define this approximate variational posterior $q(x)$ as point estimate $q(\mathbf{x})=\delta(\mathbf{x}-\eta) $\cite{graikos2022diffusion}, and then minimize:
\begin{equation}
\small
\begin{aligned}
F = \sum_t\mathbb{E}_{\epsilon\sim \mathcal{N}(0,I)}[\left \Vert \epsilon - \epsilon_{\theta}(x_t,t) \right \Vert^2_2] -\mathbb{E}_{q(\mathbf{x})}[\log{c(\eta,y)}], \\ x_t= \sqrt{\bar{\alpha}_t}\eta + \sqrt{1-\bar{\alpha}_t}\epsilon.
\end{aligned}
\label{E7}
\end{equation}
This equation optimizes $\eta$, which has the same dimensionality as data. 
We can regard this equation as directly sampling pixels using diffusion models to get a sample that satisfies the condition $y$.

Another situation is that $c(x,y)$ is a hard and non-differentiable conditional model, say a deterministic function $x=f(y)$.  
Then the gradient descent steps will be performed concerning $\mathbf{y}$ on
\begin{equation}
\small
\begin{aligned}
\sum_t\mathbb{E}_{\epsilon\sim \mathcal{N}(0,I)}[\left \Vert \epsilon - \epsilon_{\theta}(\sqrt{\bar{\alpha}_t}f(y) + \sqrt{1-\bar{\alpha}_t}\epsilon,t) \right \Vert^2_2].
\end{aligned}
\label{E8}
\end{equation}
For example, $f()$ can be a latent-variable model that takes latent $y$ as input and generates a sample $x$.
Another example is that we can also use techniques of differentiable image parameterization. 
In \cite{poole2022dreamfusion}, $y$ can be parameters of a 3D volume, and $f$ is a volumetric renderer.
We can regard this equation as sampling $y$ instead of directly sampling images using diffusion models and inputting $y$ to this conditional model. We will get a sample from the diffusion model.

Optimizing simultaneously for all $t$ makes it difficult to guide the sample toward a mode. Thus existing methods eighter anneal $t$ from high to low values \cite{graikos2022diffusion}, or random select $t$ \cite{poole2022dreamfusion}.
So the actual optimization process is slightly changed to
\begin{equation}
\small
\begin{aligned}
\mathbb{E}_{\epsilon,t}[\left \Vert \epsilon - \epsilon_{\theta}(\sqrt{\bar{\alpha}_t}f(y) + \sqrt{1-\bar{\alpha}_t}\epsilon,t) \right \Vert^2_2]
\end{aligned}
\label{E9}
\end{equation}

\section{Fine-grained control of image manipulation}
Another benefit of our approach is that the control capability of the image manipulator can be used to control the semantics more precisely. We use StyleGAN \cite{karras2019style} as the final generator of images, and the hierarchical nature of StyleGAN allows us to decompose more complex manipulations into different levels of manipulation. For example, we can adjust the manipulation effect of the image in three levels: coarse, medium, and fine. For better control, we used the most controlled StyleSpace \cite{wu2021stylespace} during the experiments. For more details, please watch the supplementary video.

\section{More visual samples}
\begin{figure*}[htbp]
    \centering
    \includegraphics[width=16cm]{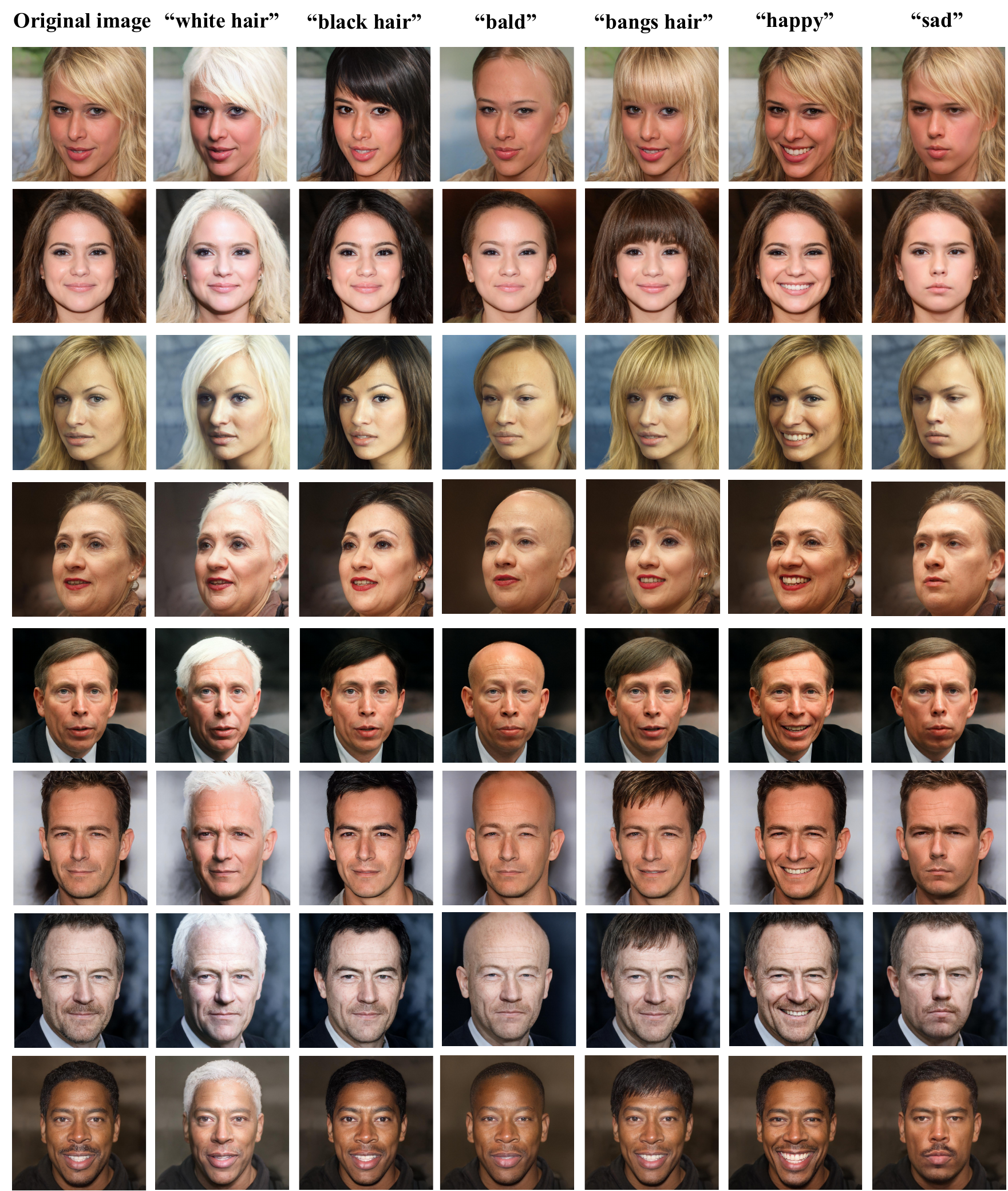}
    \caption{Additional results for SDD face manipulation, in the aspect of hair color, hairstyle, and facial expression}
    \label{sup-face1}
\end{figure*}

\begin{figure*}[htbp]
    \centering
    \includegraphics[width=16cm]{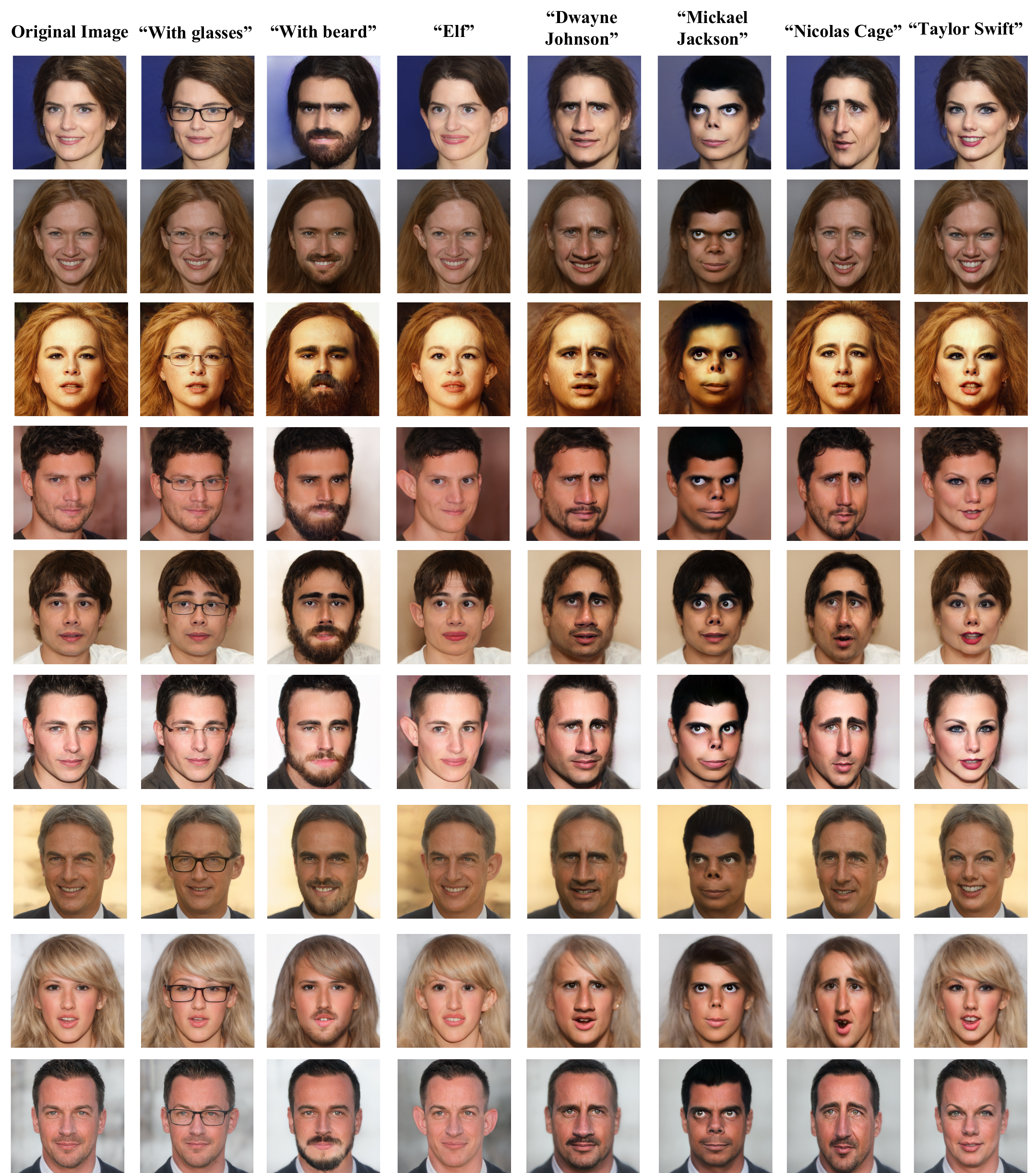}
    \caption{Additional results for SDD face manipulation, in the aspect of attributes addition and celebrities conversion}
    \label{sup-face2}
\end{figure*}

\begin{figure*}[htbp]
    \centering
    \includegraphics[width=16cm]{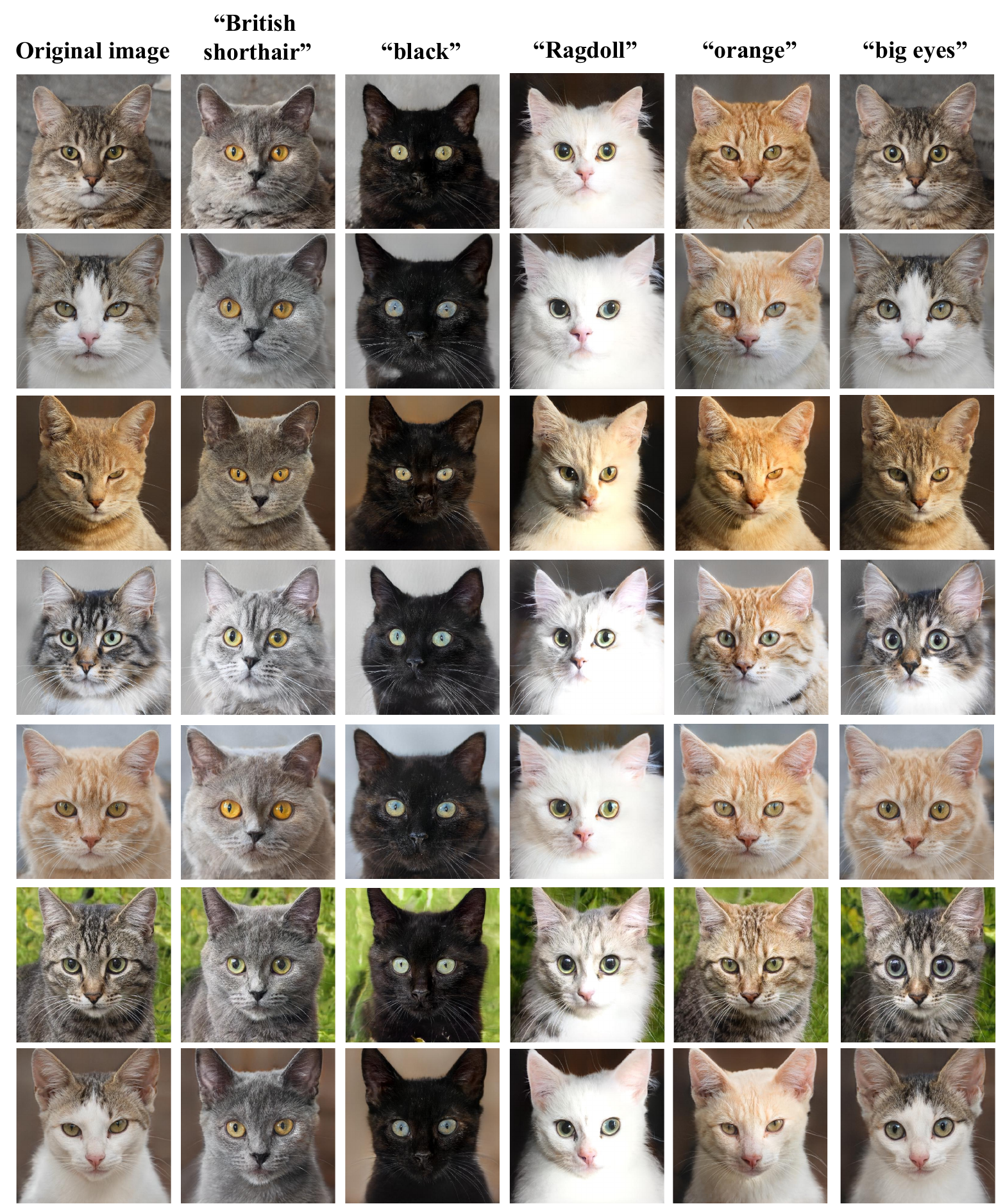}
    \caption{Additional results for SDD cat face manipulation}
    \label{sup-cat}
\end{figure*}

\begin{figure*}[htbp]
    \centering
    \includegraphics[width=16cm]{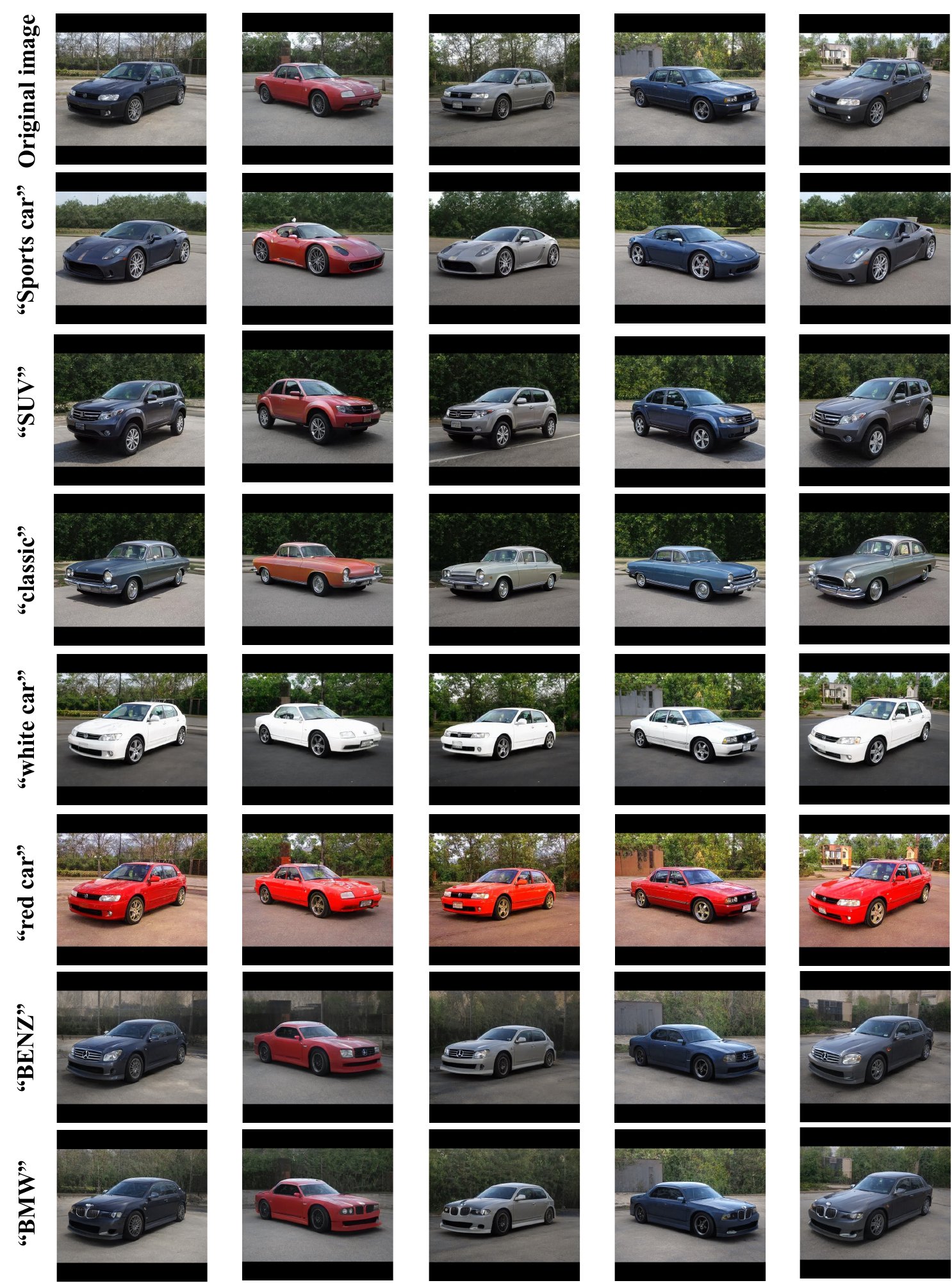}
    \caption{Additional results for SDD car manipulation}
    \label{sup-car}
\end{figure*}

In this section, we provide more visual samples (Fig. ~\ref{sup-face1}, Fig. ~\ref{sup-face2},  Fig. ~\ref{sup-cat} and Fig. ~\ref{sup-car}) from multiple domains. For different domains, the StyleGAN generator is pre-trained using different datasets \cite{liu2015deep, choi2018stargan, yu2015lsun}. 
We conduct various manipulation for the human face domain, including attribute and identity translation. Both results demonstrate the effectiveness of our methods.

\section{Video}
We summarize the analysis of selection and distillation, the fine-grained control of our method, and more visual samples in a supplementary video.


\clearpage

